\documentclass{article}                                                          
\usepackage[a4paper,total={6in,8in}]{geometry}                                                          

\usepackage{float}
\usepackage{nsfproposal}
\usepackage{tikz}                                                          
\usepackage{graphicx}
\usepackage{amssymb}
\usepackage{epstopdf}
\usepackage{graphics}
\usepackage{amsmath, amssymb}
\numberwithin{equation}{section}
\usepackage{graphicx}   
\usepackage{verbatim}   
\usepackage{color}      
\usepackage{subfig}  
\usepackage{xspace}
\usepackage{float}
\usepackage[font={footnotesize}]{caption}
\usepackage{wrapfig}
 
\DeclareMathOperator*{\argmax}{arg\,max}

\newcommand{\bff}{\mathbf{f}}
\newcommand{\bfr}{\bff^{R}}

\newcommand{\bfz}{\mathbf{z}}

\newcommand{\bfh}{\mathbf{h}}

\begin{document}
\title{Generalized Shared Control versus Classical Shared Control: Illustrative Examples}{GSC versus CSC: Illustrative Examples}
\author{Pete Trautman}

\maketitle


\section{Introduction}
\label{sec:intro}
\noindent \begin{wrapfigure}{r}{0.44\textwidth}
\includegraphics[scale=0.36]{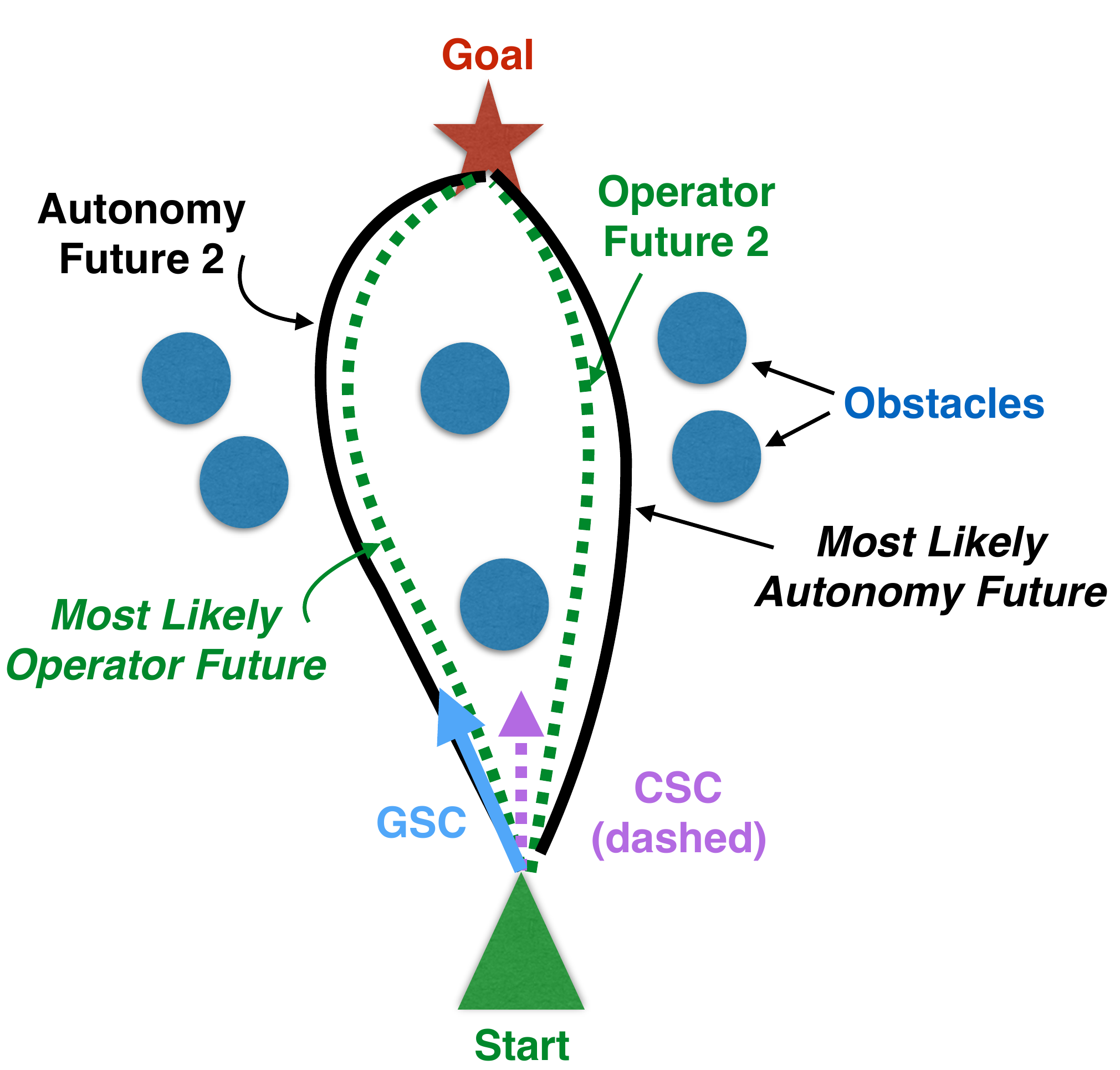}
  \caption{The autonomy (black solid lines) can take 2 possible routes through the obstacle field (thus displaying multimodality), both of which have nearly the same safety and efficiency.  The operator model (red dashed lines) indicates two potential paths forward (such multimodality might arise as a result of unreliable communications, a multitasking operator, or a startled operator).  Classical shared control (purple dashed arrow) blends the ``most likely operator future'' on the left with the ``most likely autonomy future'' on the right, resulting in an operator-disagreeable solution and a possible collision.  Generalized shared control (GSC, brown solid arrow) can blend either of the two operator futures with either of the two autonomy futures.  In this case, the \emph{data drives} GSC to blend the ``most likely operator future'' with ``autonomy future 2'', resulting in a solution that is safe and in agreement with the operator.}
  \label{fig:multimodal-operator}
  \vspace{-10pt}
\end{wrapfigure}Shared control fuses operator inputs and autonomy inputs into a single command.  However, if environmental or operator predictions are multimodal, state of the art approaches are suboptimal with respect to safety, efficiency, and operator-autonomy agreement: even under mildly challenging conditions, existing approaches can fuse two safe inputs into an \emph{unsafe} shared control~\cite{trautman-gsc}.   Multimodal conditions are common to many real world applications, such as search and rescue robots navigating disaster zones, teleoperated robots facing communication degradation, and assistive driving technologies.    In~\cite{trautman-smc-2015,trautman-gsc}, we introduced a novel approach called \emph{generalized shared control} (GSC) that simultaneously optimizes autonomy objectives (e.g., safety and efficiency) and operator-autonomy agreement under multimodal conditions; this optimality prevents such unsafe shared control.  In this paper, we describe those results in more user friendly language by using illustrations and text.

\vspace{-10pt}
\paragraph{Description of the problem}  Fusing human and machine decision making is a fundamental challenge in many areas of robotics, spanning teleoperation (search and rescue robots, UxVs, space robots, telepresence robots, and telesurgery robots) to onboard assistance (assistive wheelchairs, assistive driving, assistive medicine, and assistive orthotics/exoskeletons).  As argued in~\cite{dragan-ijrr-2013}, an approach called \emph{linear blending} (weighted averaging of the human and machine inputs) is the \emph{de-facto} decision fusion architecture in many implementations of the above application spaces.  More generally,~\cite{trautman-gsc} argues that ``human machine teaming has, for decades, been conceptualized as a \emph{function allocation} (FA) or \emph{levels of autonomy} (LOA) process: the human is suited for some tasks, while the machine is suitable for others, and as machines improve they take over duties previously assigned to humans.  A wide variety of methods---including adaptive, adjustable, blended, supervisory and mixed initiative control, implemented discretely or continuously, as potential fields, as virtual fixture interfaces, or haptic interfaces---are derivatives of FA/LOA.''  The authors go on to formalize FA/LOA under a single mathematical architecture called \emph{classical shared control} (CSC).

In~\cite{trautman-gsc}, we proved that CSC, despite its widespread usage, is only suitable as a decision fusion architecture for simple scenarios---in particular, CSC is inappropriate for scenarios displaying multimodality\footnote{Examples of ``multimodal'' scenarios include: dynamic/cluttered/responsive environments or situations where the operator only issues partial inputs (due to, for instance, unreliable networks or multitasking operators).} because it can provoke unnecessary ``disagreement'' between the operator and the autonomy.   For instance, if an operator is trying to navigate a remote platform through a dynamic arena (e.g., search and rescue robots operating in urban areas, military robots deployed into hostile environments, or even a telepresence robot trying to navigate a crowded office), CSC is nearly guaranteed to induce unnecessary operatory-autonomy disagreement.  This disagreement can, under the best circumstances, manifest as ``autonomy surprise'' (for autopilots in commercial and military aircraft, this is called mode confusion; see~\cite{rushby-mode-confusion}).  In the worst case, operator-autonomy disagreement can take a safe human command and a safe autonomy command, and generate an \emph{unsafe}  shared control (e.g., a collision).   Furthermore,~\cite{trautman-smc-2015,trautman-gsc} shows that attempts to alleviate operator-autonomy disagreement \emph{within} CSC are prone to statistical inconsistency.

\paragraph{Description of the solution} These results mandate that in order to make progress in shared control, CSC must be replaced.  Fortunately, the results in~\cite{trautman-gsc} provide a fundamentally different approach  called \emph{generalized shared control} (GSC).  GSC optimizes human-robot agreement and individual agent intent in multimodal situations. Because the GSC framework is designed to optimally share control through operator or environmental multimodality, our results can be applied to a wide range of problems.  Consider unreliable communications between the operator and the autonomy, so the autonomy is only receiving partial information from the operator. This partial information will cause the amount of uncertainty in the operator's \emph{predicted} intent to increase. Mathematically, this uncertainty can manifest as a multimodal distribution (a model that forecasts multiple different futures), such as in Figure~\ref{fig:multimodal-operator}. This multimodality can instigate disagreement in a CSC framework; GSC, on the other hand, is designed to mitigate disagreement while preserving the objectives of the autonomy under these types of scenarios.  

\section{Technical Description}
\label{sec:tech-description}

We adopt a fundamentally different strategy in formulating shared control.   We start from the insight that shared control should capture the interdependent relationships between the human operator, the autonomy, and the environment; that possible future configurations can effect decisions of the shared controller now; and that predictive uncertainty about the operator and the environment is significant.  We thus use a joint probability distribution over the future \emph{trajectories} of the autonomy, the operator, and the environment:  probability distributions are well suited to model uncertainty; joint distributions capture the interdependencies between the operator, autonomy, and environment; and by using a joint distribution over trajectories, we can model how future configurations can affect decision making now.  As was described in~\cite{trautmanicra2013}, once we have such a ``cooperation distribution'', we can naturally interpret the shared control as the joint maximum value, since this statistic optimally balances the competing interests of the the human operator, the dynamic environment, and the robot.


To fully explain the novelty of our approach, we begin by defining linear blending:
\begin{align}
\label{eq:lin-blend}
u^s_{LB}(t) = K_hu^h_t + K_Ru^R_{t+1},
\end{align}
where, at time $t$, $u^s_{LB}(t)$ is the linearly blended shared control command sent to the platform actuators, $u^h_t$ is the human operator input (joystick deflections, keyboard inputs, etc.), $u^R_{t+1}$ is the next autonomy command,  and $K_h, K_R$ are the operator and autonomy \emph{weighting factors}, respectively, and can change at each time step.  Shared control path planning researchers have widely adopted Equation~\ref{eq:lin-blend} as the \emph{de-facto} standard protocol, as extensively argued in \cite{dragan-ijrr-2013, draganrss2012} (we refer the reader to the prior work section of~\cite{dragan-ijrr-2013} for a definitive bibliography on shared control and its relationship to linear blending).  In particular, linear blending has enjoyed wide adoption in the assistive wheelchair community (\cite{carlson-smc-2012,wang-adaptive-shared-control, lopes-embs-2010, urdiales-autonrobots-2011,yu-autonrobots-2003,peinado-icra-2011, urdiales-nsre-2013,inigo-blasco-isrrobotik-2014}).  Additionally, the work of~\cite{poncela-smc-2009} and~\cite{wang-ras-2005} advocates for the broad adoption of a linear arbitration step in shared control.  In~\cite{trautman-gsc}, the authors cast a much broader net, capturing many standard approaches to ``decision fusion'' (adaptive, adjustable, blended, supervisory and mixed initiative control, implemented discretely or continuously, as potential fields, as virtual fixture interfaces, or haptic interfaces) and ultimately function allocation and the levels of autonomy paradigm as special cases of what they call classical shared control.


For completeness, we briefly describe our approach to fusing decision making, GSC.  We begin by introducing the following notation: $\bfz^h_{t} \doteq u^h_t,$  (that is, we treat operator inputs as \emph{measurements} of the operator \emph{trajectory}, $\bfh \colon t\in\mathbb R \to \mathcal X$, where $\mathcal X$ is the action space).  Similarly, we define measurements $\bfz^R_{1:t}$ of the robot trajectory $\bfr\colon t\in\mathbb R \to \mathcal X$ and measurements $\bfz^i_{1:t}$ of the $i$'th static or dynamic obstacle trajectory $\bff^i\colon t\in\mathbb R \to \mathcal X$.  We thus work in the space of distributions over the operator function $\bfh$, autonomy function $\bfr$, and ``environment'' function $\bff = (\bff^1, \ldots, \bff^{n_t})$, measured through $\bfz^f_{1:t}$.   The integer $n_t$ is the number of agents in the environment at time $t$.  We thus define the generalized shared control input ($u^s_{GSC}$) as
\begin{align}
\label{eq:probabilistic-shared-control}
\vspace{-3mm}
u^s_{GSC}(t) &= \bff^{R^*}_{t+1}\nonumber \\
 (\bfh,\bff^{R},\bff)^* &=\argmax_{\bfh, \bfr,\bff} p(\bfh, \bfr,\bff \mid \bfz^h_{1:t}, \bfz^R_{1:t},\bfz^f_{1:t}).
 \vspace{-2mm}
\end{align}
That is, generalized shared control is the most likely \emph{joint} hypothesis $(\bfh,\bff^{R},\bff)^*$ of the operator, robot, and environment (or, GSC balances the interdependent needs of the operator, robot and environment).  This approach is sometimes called ``planning as inference'' (see~\cite{toussaintrss2012}): whereas in optimal control theory an explicit cost function is optimized, with GSC the joint distribution implicitly captures costs and goals.  That is, trajectories with higher probability can be interpreted as more optimal (or lower cost).  This idea is explained in detail in~\cite{trautmaniros,trautmanicra2013}.

We point out that the full joint distribution, $p(\bfh, \bfr,\bff \mid \bfz^h_{1:t}, \bfz^R_{1:t},\bfz^f_{1:t})$, is not known in advance.  In previous work on fully autonomous navigation in crowds (where the autonomy $\bfr$ and the crowd $\bff$ was present, but the operator, $\bfh$, was not), we explored an approximation to the full joint which preserved the dependency structure between the autonomy and the crowd through an interaction function between $\bfr$ and $\bff$ (see~\cite{trautmaniros, trautmanicra2013}); this approach showed 3-fold improvement in safety and efficiency over nearly all other existing navigation algorithms during extensive real world testing.  

For GSC, we extend this fully autonomous model to include a) a model of the operator and b) an interaction function between the autonomy and the operator.  Importantly, \emph{the theorems paraphrased in Section~\ref{sec:paraphrase} hold for our model of the full joint distribution.}

\subsection{Theoretical Properties of GSC}
\label{sec:paraphrase}
 In~\cite{trautman-smc-2015,trautman-gsc}, we prove four important theorems about how our approach compares to the state of art in shared control (CSC).  We paraphrase those theorems here:
\begin{enumerate}
\item Linear blending is a highly restrictive special case of GSC.  In particular, linear blending assumes
\begin{enumerate}
\item the operator will only do one thing in the future and
\item the autonomy will only do one thing in the future.
\end{enumerate} 
This leads to a an extremely brittle decision fusion architecture;  we accordingly prove that
\item Linear blending is jointly suboptimal with respect to the autonomy's objectives (e.g., safety and efficiency) and operator-autonomy agreement.  For instance, even in \emph{static} cluttered scenarios, linear blending can fail under trivial circumstances: it can take a safe input from the operator, a safe input from the autonomy, and blend the two into a collision.  In other words, linear blending is not only just a little suboptimal---sometimes, it is \emph{catastrophically} suboptimal.  Worse still, it can be very hard to predict \emph{when} a catastrophic linear blend will occur.
\item CSC fails to optimize human-robot agreement and intent if intention ambiguity is present.
\item GSC is optimal with respect to respect to the autonomy's objectives (e.g., safety and efficiency) and operator-autonomy agreement.  
\end{enumerate}
As we explained in Section~\ref{sec:intro}, the illustrative examples in Section~\ref{sec:demo} are directly interpretable as special cases of jointly optimizing over operator-autonomy agreement, safety, and efficiency.  Our theorems indicate that not only will CSC \emph{not} work, but that GSC will work as well as \emph{any other} possible approach.

\newpage
\subsection{Illustrative Examples of GSC versus CSC}
\label{sec:demo}
\subsubsection{Remote Teleoperation over Unreliable Networks} 
\label{sec:teleop-demo}
\begin{wrapfigure}{r}{0.39\textwidth}
 \vspace{-10pt}
\includegraphics[scale=0.32]{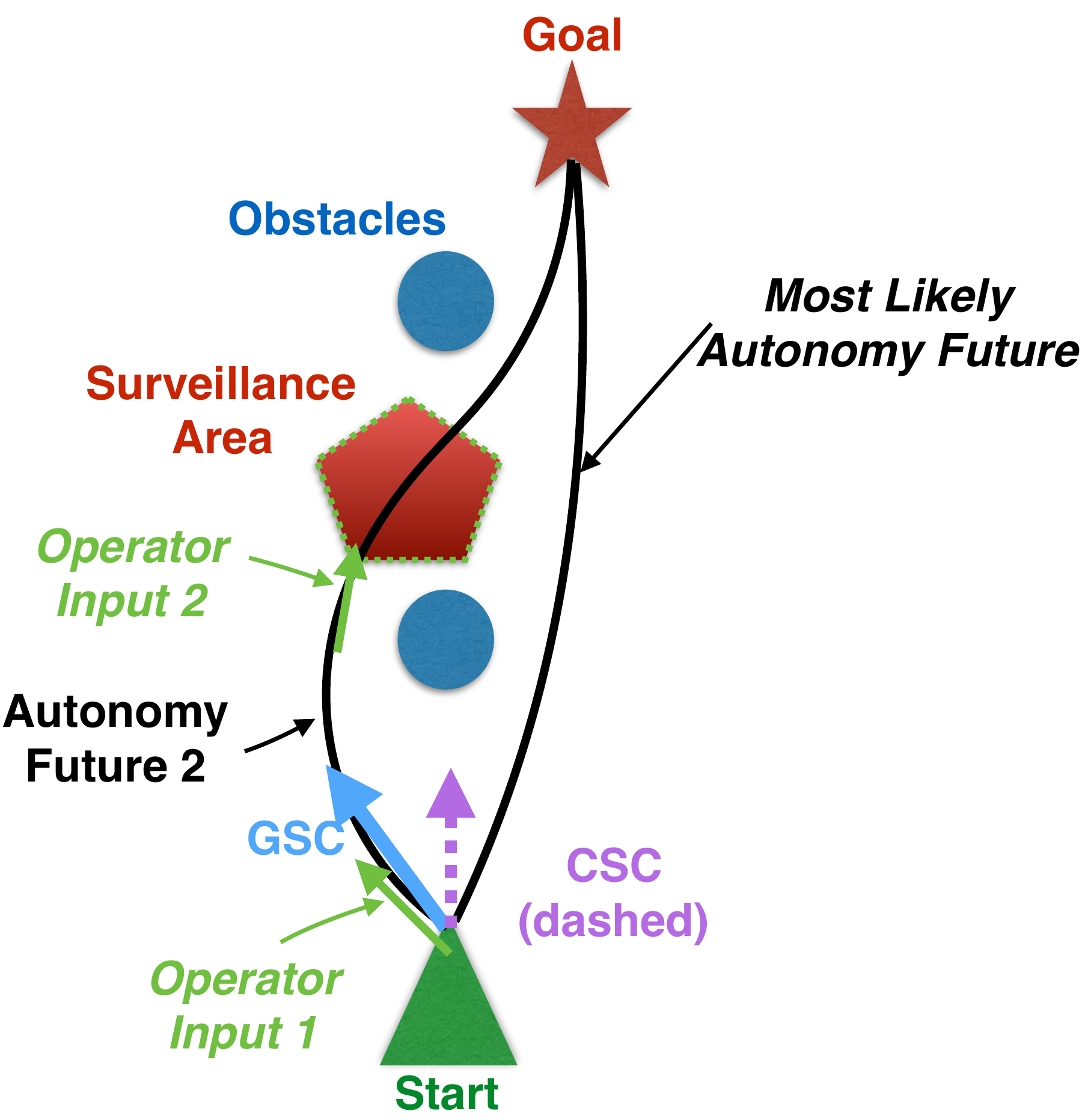}
  \caption{Lossy communications: operator is instructed to cover the surveillance area; communications are then lost (the autonomy does not have a global map, and so cannot be directed to fly over specified regions).  The rightmost path is the safest and most efficient.  Linear blending, without operator direction, follows this path, missing the surveillance area.  Operator input 1 causes GSC to reason over autonomy future 2, and so GSC is positioned to pass over the surveillance area.}
  \label{fig:lossy-example}
  \vspace{-10pt}
\end{wrapfigure}
\begin{figure}[h!]
  \centering
  \subfloat[At $t-1$, operator sees that upper four obstacles might move.  Input to go right is not received by autonomy until $t$, so confidence in operator is low.  Confidence in centerline autonomy path is high, so linear blending proceeds through the middle. GSC reasons over suboptimal path to right. ]{\includegraphics[scale=0.32]{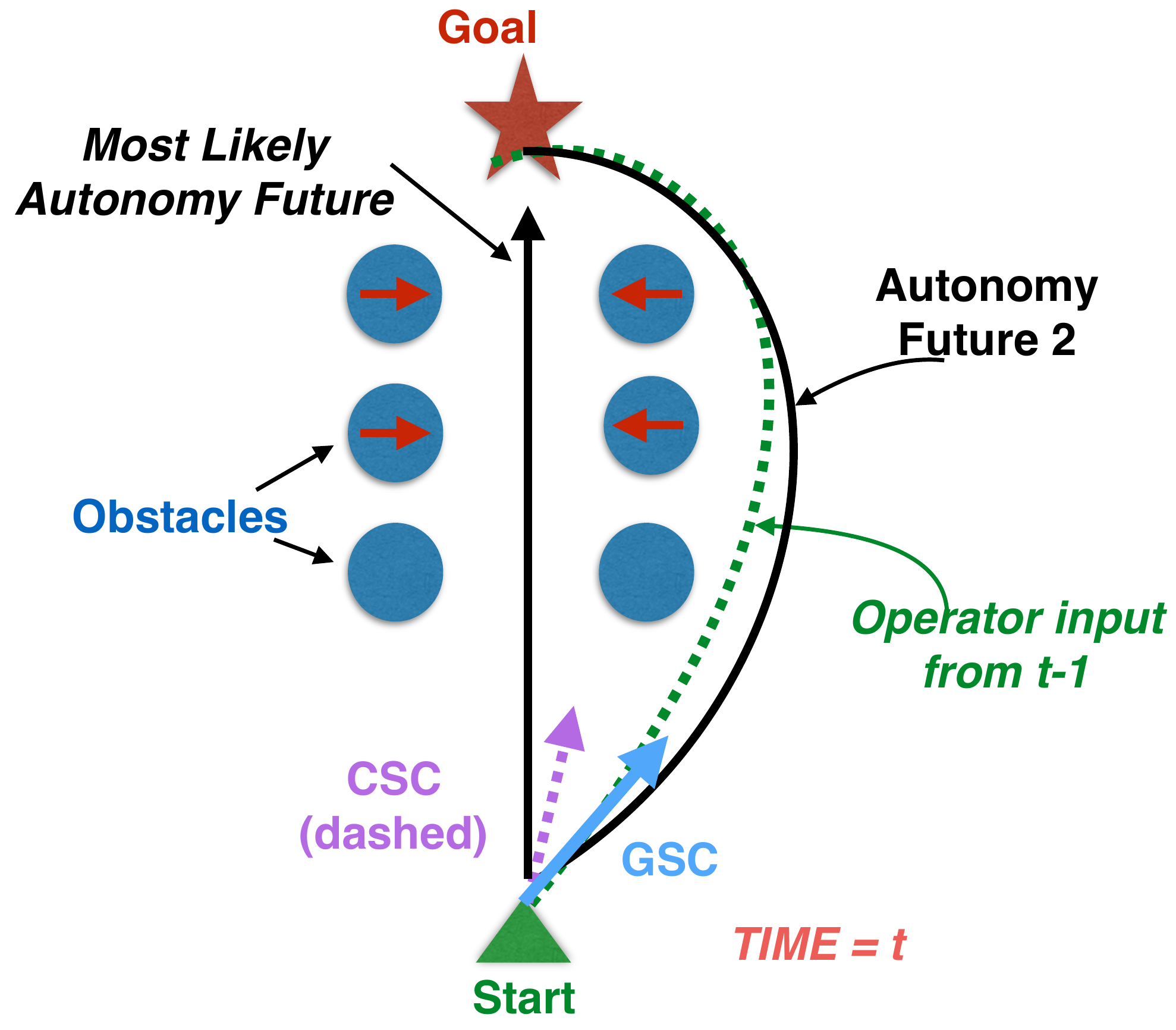}
  \label{fig:challenge1a}}
  \hfill
  \subfloat[By time $t+1$, obstacles have moved into robot's field of view, and the most likely autonomy future has snapped to the right.  Linear blending has to take an evasive maneuver to avoid collision.  GSC is well positioned because it is able to properly weight inputs from the past and reason over suboptimal paths.]{\includegraphics[scale=0.32]{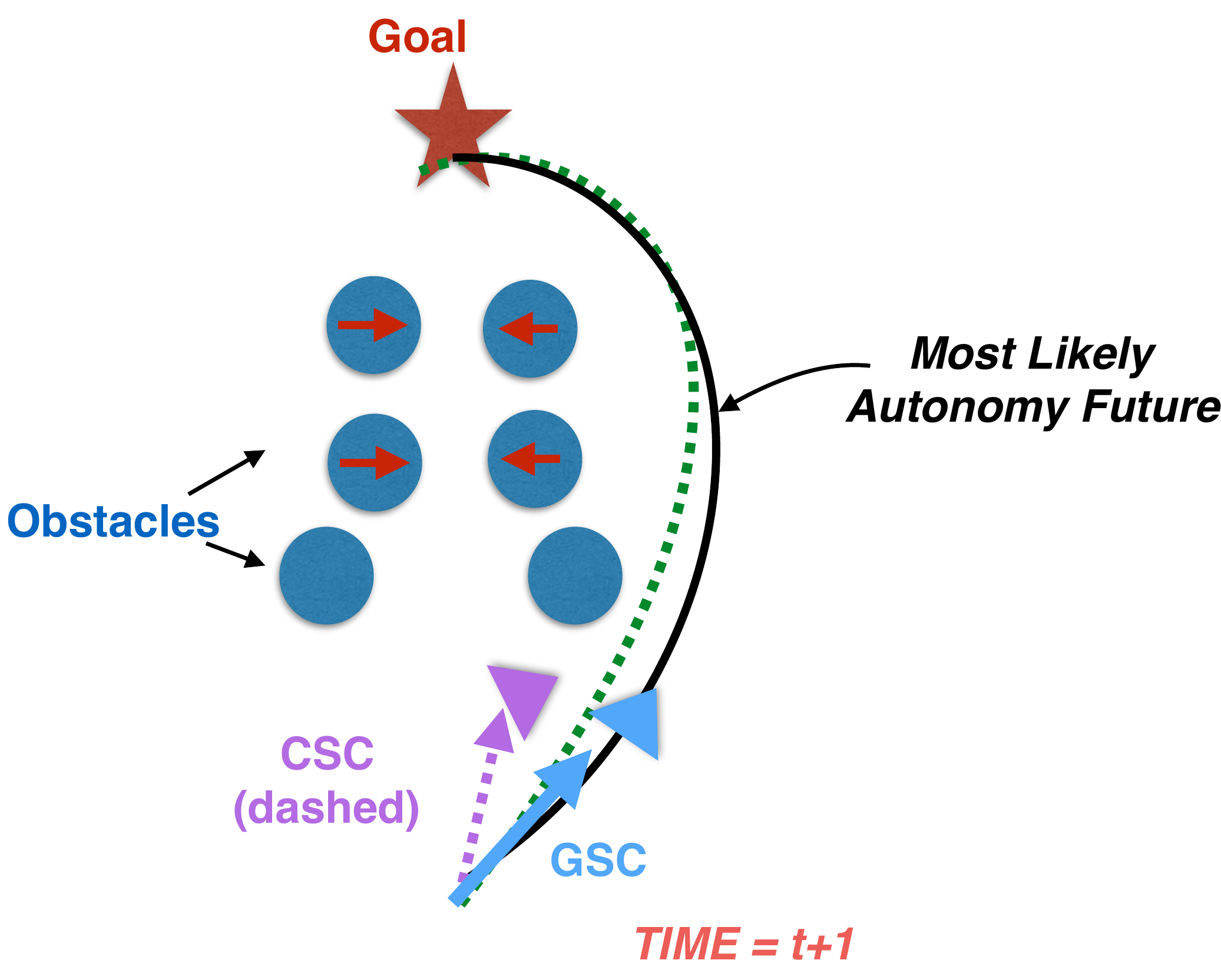}
  \label{fig:challenge1b}}
  \caption{Challenge scenario for laggy communication experiment.}
  \label{fig:challenge1}
\end{figure}

By unreliable communications we mean 1) lossy, 2) laggy, and 3) noise corrupted operator input data.  Our formulation in~\cite{trautman-gsc,trautman-smc-2015} provides a natural mechanism to handle such data sources: a predictive probabilistic model of the operator's intention (also described in~\cite{trautman-4-challenges}).  Further, our teaming framework (GSC) is designed to optimally disambiguate operator models corrupted by these kind of data errors because it reasons over \emph{multimodal} autonomy models (see Figures~\ref{fig:lossy-example} and~\ref{fig:challenge1}).  In contrast, linear blending does not carry probabilistic models and is fragile to multimodality, and so linear blending is extremely brittle to communication failures.  We thus consider the following thought experiments:
 
\noindent 1.  Lossy communications: we simulate lossy networks by randomly dropping inputs sent from the operator.  We assume that the remote operator has a fixed route in mind, but has trouble communicating this route because of communication drops.  Thus our metric of performance will be how close the remote platform stays to the operator's intended (but only partially known) route.   In Figure~\ref{fig:lossy-example}, we illustrate a scenario where GSC outperforms linear blending.

\noindent 2.  Laggy communications: what happens when an operator input is delivered to the remote platform, but is delayed by some known  amount of time?  How should that data be interpreted (see the challenge scenario below and Figure~\ref{fig:challenge1})?  Existing frameworks typically discard this information, since, without a probabilistic operator model, it is not clear how to interpret time delayed information.  Our predictive probabilistic operator model extrapolates inputs from the past and our decision fusion architecture is better suited to disambiguate the effects of laggy information.

\paragraph{Challenge Scenario: When the Operator Notices Danger Before the Autonomy}
\label{para:challenge-comm}

We present a challenge scenario for a laggy communications situation between the autonomy and the platform in Figure~\ref{fig:challenge1}.  At time $t-1$, the operator notices a passageway is closing through the middle of the obstacle field, and so directs the autonomy to go right; this input is not received by the autonomy until time $t$.  However, the autonomy's field of view is occluded by the foremost obstacles and so the autonomy thinks the path through the middle is the most efficient.  Communications then drop out.  Because the operator information is 1 second old, the operator model has low confidence in the operator input.  Thus, linear blending has high confidence in the autonomy trajectory through the middle and low confidence in the operator input, and so produces a shared control that is biased towards the autonomy path.  
\emph{Importantly, because CSC only reasons over a finite set of  autonomy futures, its ability to weight alternatives is compromised.}  GSC has the same low confidence in the operator, but is able to weight against the lower probability trajectory to the right. \emph{Since GSC can reason over many possible operator-autonomy combinations, it can find the shared control which optimizes safety, efficiency, and operator-autonomy agreement.} The autonomy does not notice the moving obstacles until a few moments into the future, at which point the most  likely autonomy future snaps to the right (for both linear blending and GSC, since they are both on the right side of the midline), as shown in Figure~\ref{fig:challenge1b}.  When this occurs, linear blending has to take an evasive maneuver to recover the most likely autonomy future, while GSC is already very close to the safe trajectory.

\subsubsection{Shared Control for Task Handoffs}  
\begin{figure}[t!]
  \centering
\subfloat[A distracted human does not see the dashed red obstacle (but the robot does) so tries to proceed directly forward.  CSC is forced to reason over the human input and the autonomy input, which potentially places it in an inevitable collision state.  GSC does not maintain probability in regions that it knows are occupied by obstacles, and so blends closely to the most likely autonomy future.  \emph{N.b.: CSC can include a collision safeguard that prevents these situations.  In Figure~\ref{fig:context-cue}, we illustrate how such a protocol can have unintended consequences.}]{\includegraphics[scale=0.3]{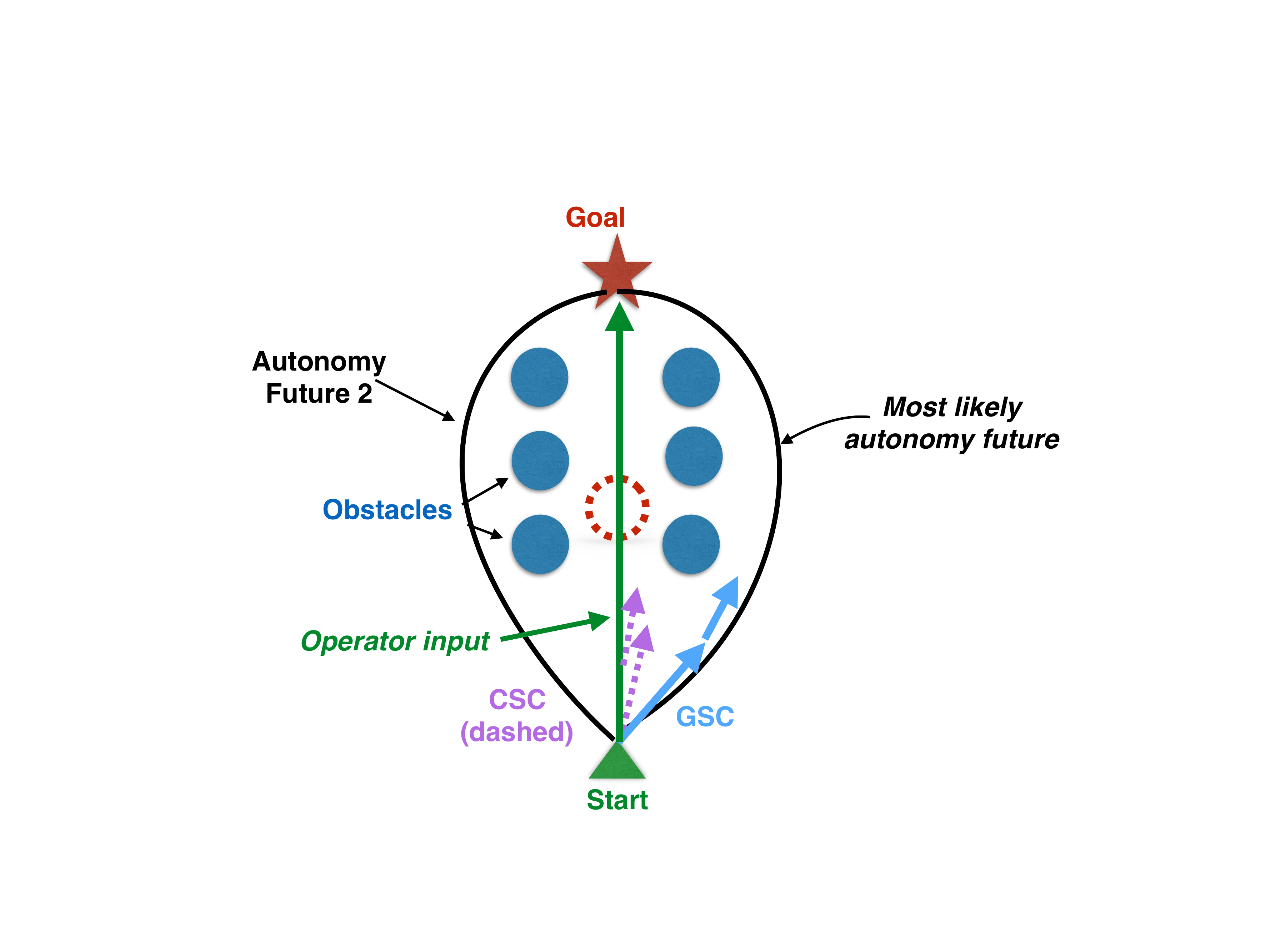}
  \label{fig:human-wrong}}
  \hfill
  \subfloat[The operator hears ``elevator coming!'' and tells the autonomy to proceed directly to the service elevator.  The autonomy is unaware that the hallway is about to empty; the optimal robot path is to creep along the wall.  \emph{Linear blending has a collision override}, so the human's input is ignored, and linear blending goes right.  GSC blends over a low probability mode through the crowd (maximizing operator-autonomy agreement).  As crowd empties, GSC moves directly to the goal; linear blending is caught between the pedestrians and the elevator.]{\includegraphics[scale=0.3]{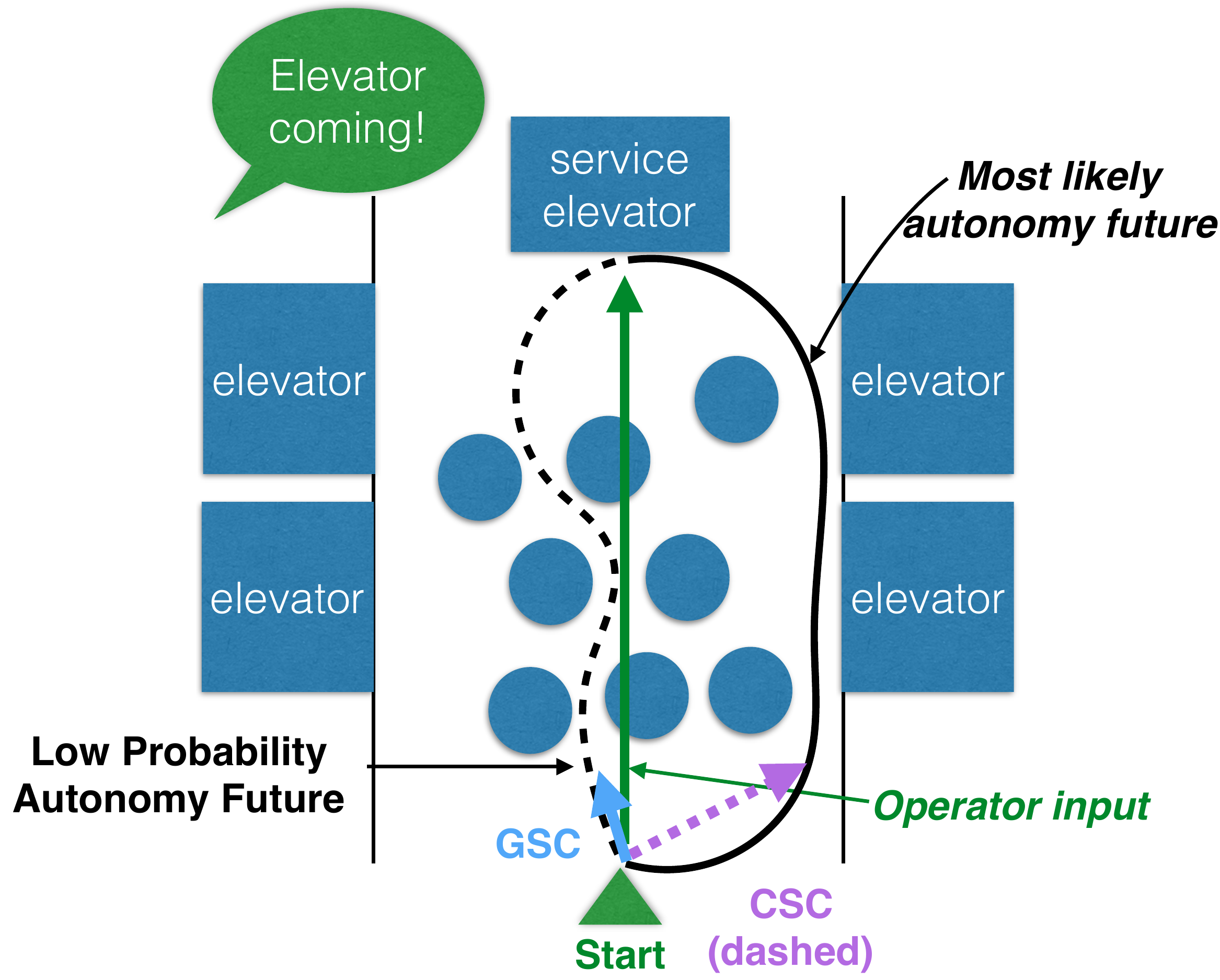}
  \label{fig:context-cue}}
   \caption{Challenge scenarios for task handoffs.  If linear blending is designed with a collision safeguard (e.g., to protect a distracted operator,~Figure~\ref{fig:human-wrong}), it is unable to leverage human insight from subtle environmental cues (Figure~\ref{fig:context-cue}).  Because GSC is \emph{data-driven}, it can handle task handoffs in either direction. }
 \label{fig:challenge2}
\end{figure}

Operators are often distracted from the task of navigating the platform for legitimate reasons: a search and rescue operator might be looking for survivors buried in rubble, a UAV operator might be reviewing new intelligence about insurgent activity in his surveillance area, or a commercial telepresence user might be engaged in a conversation while navigating 
down a crowded hallway.   A primary motivation for shared control is precisely these situations: the autonomy should gracefully take over tasks when the operator cannot.  Similarly, the limitations of artificial intelligence mean that the autonomy might miss important contextual cues that a human operator can often exploit; shared control architectures should leverage human instincts that exceed the capability of autonomous functionality.  Importantly, GSC is well suited for these types of task handoffs.  CSC, on the other hand, has fundamental deficiencies that make it unable to handle handoffs in \emph{both} directions (where either the operator or the autonomy excels, see the challenge scenario below and Figure~\ref{fig:challenge2}). For a number of scenarios, the distracted operator can be mathematically characterized by extended periods during which the operator provides no input to the platform, or short periods during which incorrect input is provided to the platform.  Likewise, scenarios where the autonomy misses a contextual cue that the operator sees can be characterized by an intermediate level of operator-autonomy disagreement.  

\paragraph{Challenge Scenario: Data Driven Versus Heuristic Approaches} A distracted operator can miss a danger that the robot sees (e.g., a short obstacle that gets picked up by a laser scanner, or turbulence for an air vehicle), and thus the operator can direct the robot towards the danger.  In these situations, GSC will guide the operator around the danger, as in Figure~\ref{fig:human-wrong}, in a data driven way: with GSC, the shared control is based on the existing data, rather than hand tuned heuristics.  Standard implementations of linear blending do not protect against such situations (as illustrated in Figure~\ref{fig:human-wrong}), but can be \emph{fixed} by using collision safeguards.  For Figure~\ref{fig:human-wrong}, the safeguard is straightforward: ignore unsafe human inputs.  However, this safeguard can have unintended consequences. In Figure~\ref{fig:context-cue}, a crowd of pedestrians waits for elevator doors to open.  At the top of the figure is a robot service elevator.  The operator hears the ``elevator coming!'' signal, and knows that a direct path to the service elevator will soon open.  Linear blending, with the collision safeguard in place, ignores the operator, and takes an evasive maneuver towards the edge of the hallway.  GSC, on the other hand, maintains low probability modes \emph{through} the crowd, and because of the operator's instruction, follows the lead of the operator. The GSC robot thus proceeds directly to the service elevator, while the linear blending robot is caught near the edge of the crowd entering the elevator.  Linear blending has both overridden the operator and ended up in a precarious situation.  
Figures~\ref{fig:human-wrong} and~\ref{fig:context-cue} illustrate an important quality of GSC (that is not present in linear blending): it is \emph{data driven} (meaning the data drives the behavior of the shared control), and as such, requires minimal hand tuning for proper performance.  In real world situations, where potential edge cases are infinite, hand tuning heuristics (e.g., collision safeguards) is not a scaleable or reliable strategy.

\subsubsection{Assistive driving: the Startled Driver and the Overcautious Autonomy.}
\begin{figure}[h!]
  \centering
\subfloat[Illustration of the startled driver: the autonomy is  safely proceeding to an intersection, when the operator looks up and notices cars and pedestrians, and jerks the wheel (green).  GSC has no autonomy modes crossing into oncoming traffic, and so the operator input is ignored.  Linear blending averages the operator and autonomy input into an unsafe movement.  If linear blending has a collision guard, then the operator will be ignored; however, as discussed in~Figure~\ref{fig:congestion}, safeguards can force the robot to freeze in place in congested scenarios.]{\includegraphics[scale=0.32]{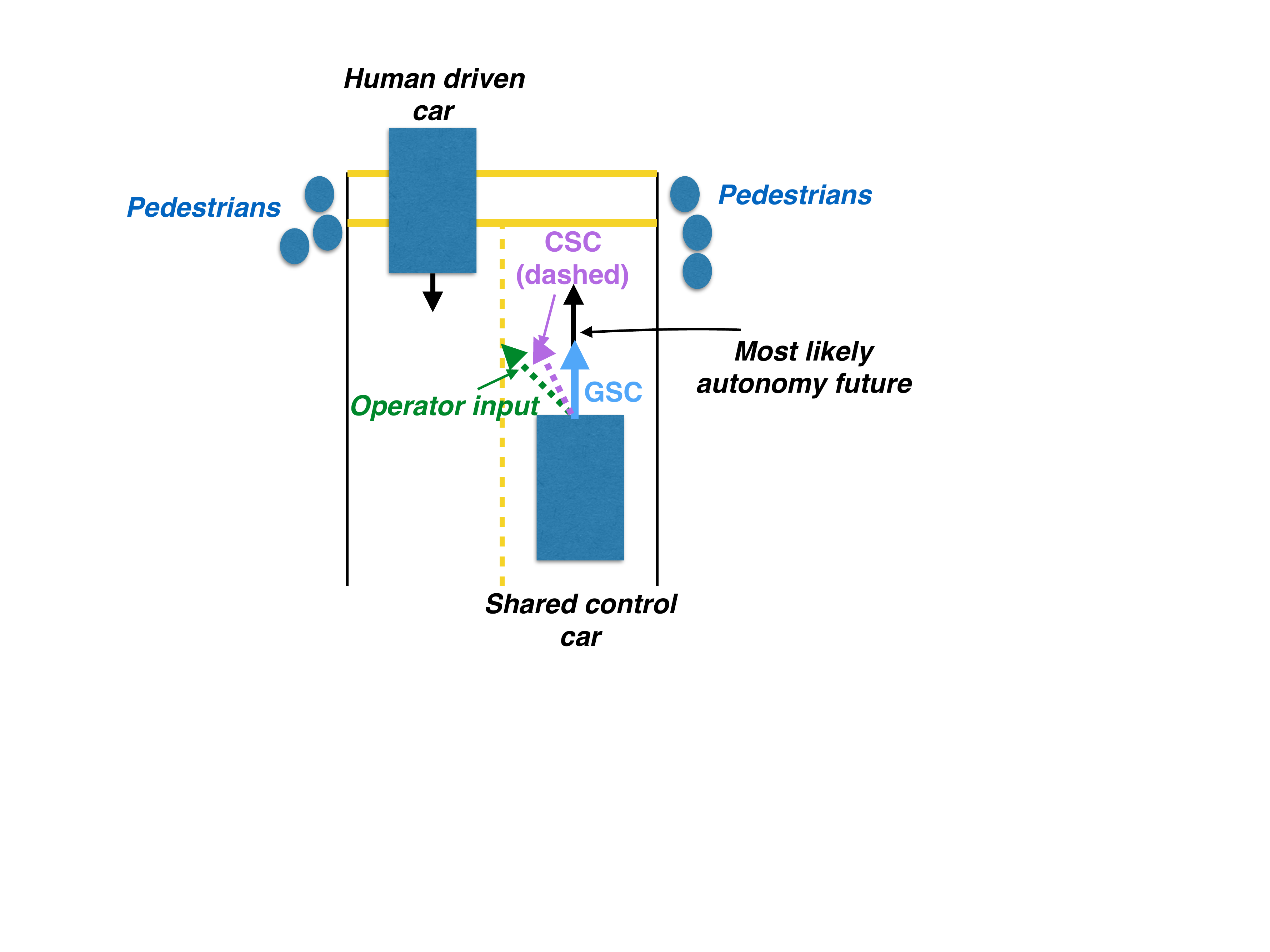}
  \label{fig:startled-driver}}
  \hfill
  \subfloat[A standing problem for fully autonomous cars: how aggressive should the autonomy be?  Here, a shared control car needs to merge with traffic.  The autonomy only has low probability modes of safe merging, and will thus never merge on its own.  Humans can spot safe merging opportunities (e.g., via eye contact with a driver).  Linear blending cannot merge with a collision guard; without a guard, it crashes (see~Figure~\ref{fig:startled-driver}).  GSC combines machine precision and human guidance to produce a safe merge. ]{\includegraphics[scale=0.32]{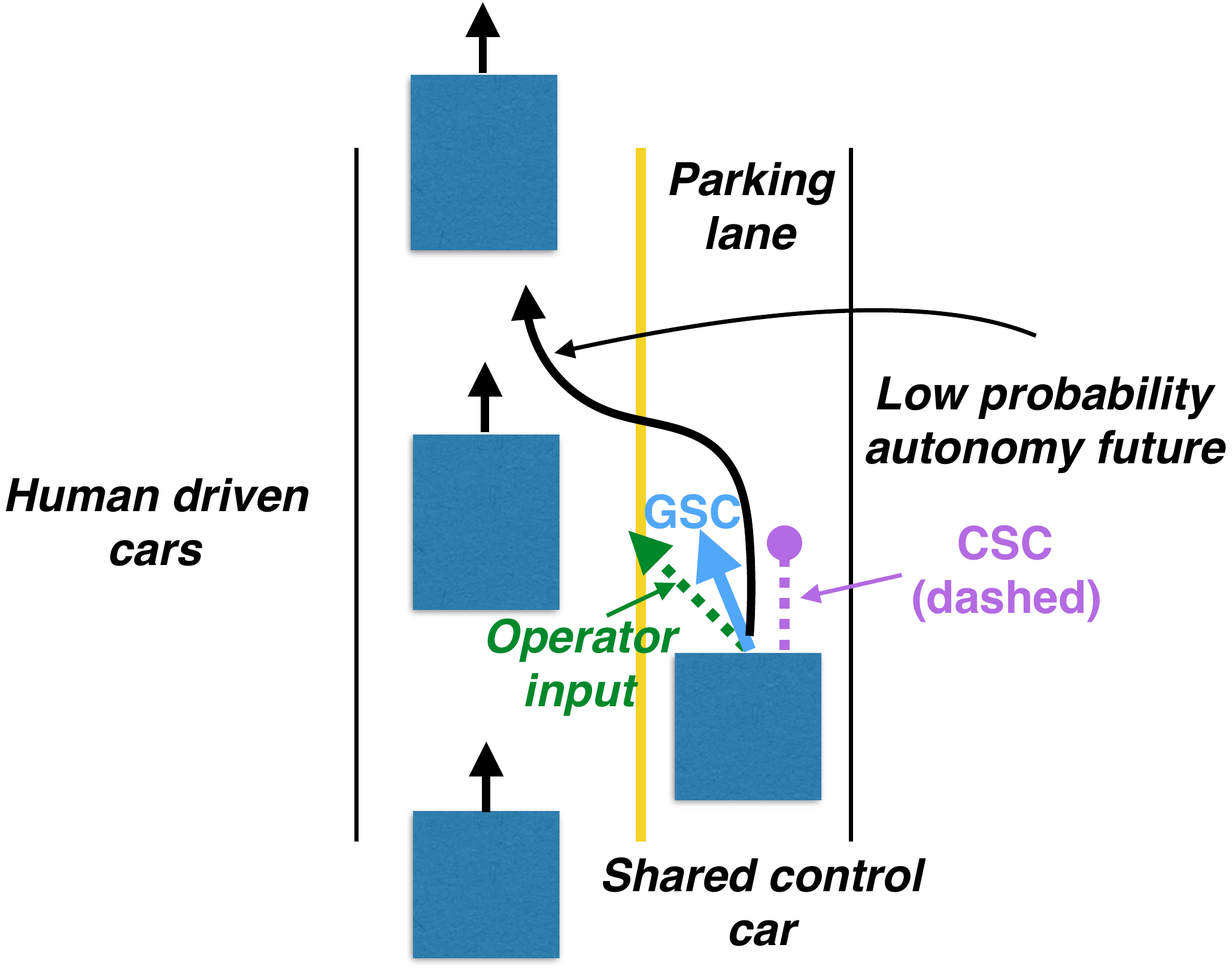}
  \label{fig:congestion}}
  \caption{Challenge scenarios for assistive driving: a startled input causes linear blending \emph{without} collision guards to crash (Figure~\ref{fig:startled-driver}); crashes are prevented with a collision guard.  In Figure~\ref{fig:congestion}, linear blending \emph{with} a collision guard cannot negotiate heavy traffic situations.  CSC thus cannot be used for assistive driving.}
  \label{fig:challenge3}
\end{figure}

\noindent  For shared control in autonomous cars, one of the most challenging situations is the inattentive driver (texting, talking on the phone, reading a book) who suddenly looks up, decides the autonomy is headed towards a catastrophe (even though the autonomy has the car under control), and jerks the wheel or slams on the brakes.  
For switching control, once the driver grabs the wheel complete control is handed over to the the human---suddenly jerking the steering wheel or slamming on the brakes can have catastrophic consequences (imagine this happening on a highway or a narrow street crowded with pedestrians).  GSC handles this situation in a principled manner: since the autonomy has no forward modes crossing into oncoming traffic, the operator is ignored, and the car proceeds safely.  With linear blending, if a collision guard is in place (i.e., ignore unsafe operator inputs), the car proceeds safely; without a collision guard, linear blending produces an unsafe trajectory (Figure~\ref{fig:startled-driver}).

So why not just use linear blending with a collision guard, since it is a simpler approach than GSC?  In Figure~\ref{fig:congestion}, we illustrate another standing challenge for self driving cars (both full and shared autonomy): merging into heavy traffic.  Because GSC reasons over low probability modes, it is able to follow the human operator's lead when the person spots a hole in traffic.  However, if linear blending has a collision guard in place (to prevent accidents in scenarios like that illustrated in Figure~\ref{fig:startled-driver}), then it stays frozen in place; because the autonomy cannot find a safe enough route, linear blending \emph{always} overrides the operator (and since the optimal \emph{autonomy} path is to stay in place, the CSC action is to not move the car).  Removing the collision guard makes the shared control dangerous.  Situations like these demonstrate why CSC is unsuitable for real world applications.

\subsection{Use Cases and Beneficiaries of Generalized Shared Control}
\label{sec:use-cases}
In addition to the use cases described in Section~\ref{sec:demo}, we describe other potential use cases for GSC.

\begin{enumerate}
\item Search and rescue: an important aspect of search and rescue is that a first responder might be looking for contextual clues about, e.g.,  where a person is buried under rubble.  Often, robot perception is not very good at deducing context.  For instance, a first responder might spot the tip of a buried hand, while a robot may have trouble connecting a random body part to a buried human.  Thus, a search and rescue robot needs to be able to relieve the first responder of the cognitive burden of navigating the robot (so the human can look for contextual clues), but the robot still needs to be able to navigate to the correct locations given \emph{partial} operator direction (compare with Section~\ref{sec:teleop-demo}).

\item Commercial/military autopilot technology.  A longstanding problem here is mode confusion:  the human pilot hands off control to the autopilot, and the autopilot does something that the human does not understand.  Mode confusion is potentially dangerous if the pilot cannot make sense of the autopilot's actions, and so understanding the provenance of mode confusion is important.  We note that \emph{disagreement} (such as illustrated in Figure~\ref{fig:two-mode-autonomy}, in which the operator wants to go left and the autonomy wants to go right) logically precedes mode confusion---otherwise, the pilot agrees with what the autopilot is doing, and thus is not confused.  \begin{figure}[h]
  \centering
\includegraphics[scale=0.33]{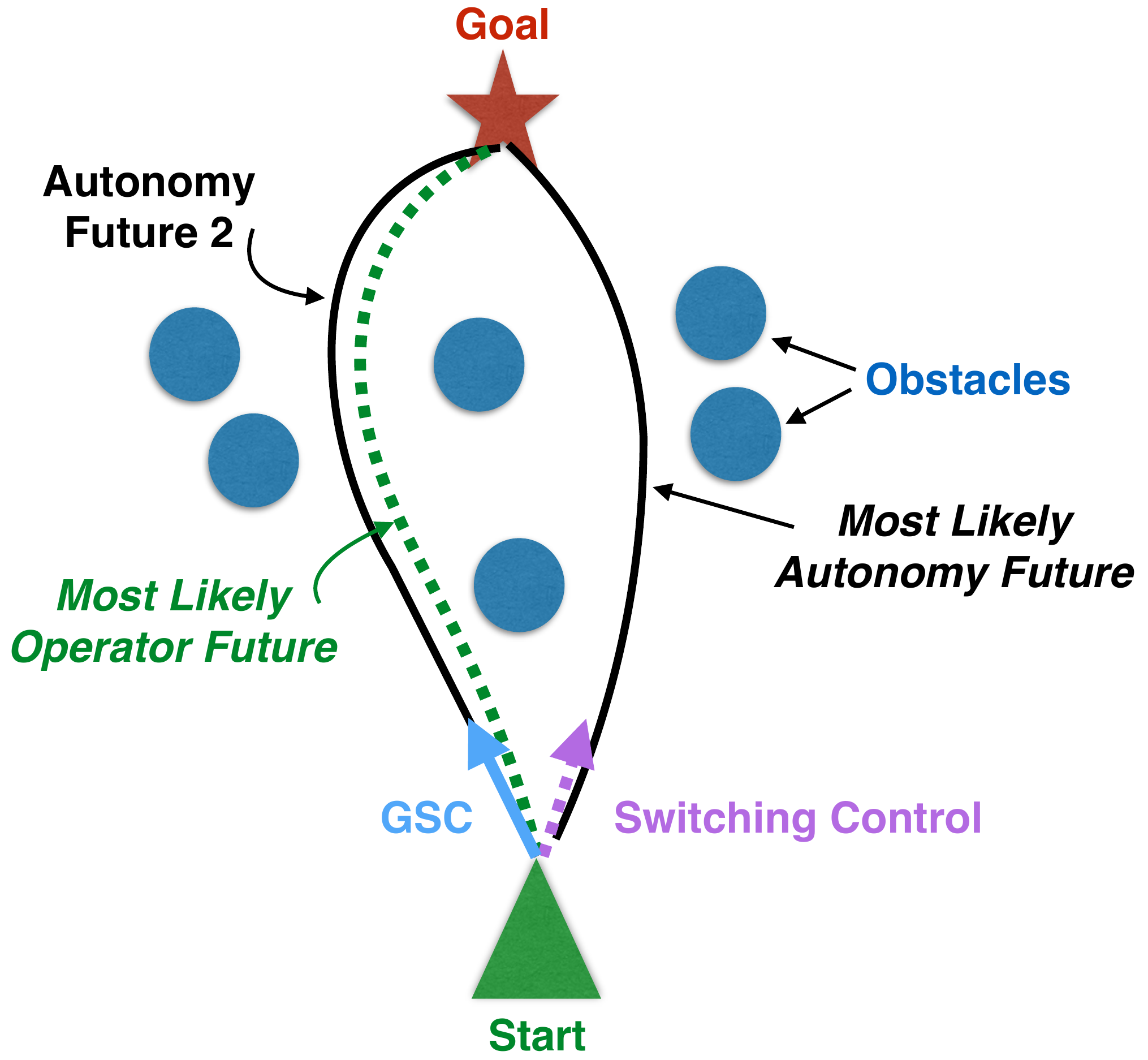}
  \caption{In this scenario, the autonomy can take two possible trajectories (black solid), the right one being slightly safer and more efficient.  However, the pilot clearly indicates a desire to go left (red dashed).  Switching control (purple dashed arrow) must choose between two irreconcilable actions (going left versus going right).  GSC (light blue arrow), on the other hand, is able to reason over the slightly suboptimal leftmost trajectory, which blends much more naturally with the operator trajectory. \emph{ Imagine trying to explain the decision making of linear blending/switching control versus GSC for this situation}.}
  \label{fig:two-mode-autonomy}
\end{figure}We further note that existing avionic suites share control via switching: either the autopilot or the pilot is controlling the aircraft.  As noted earlier, switching control is a special case of linear blending, and as such, is vulnerable to unnecessary disagreement.   Could such disagreement be the (partial) cause of mode confusion?  It seems that if disagreement could be maximally mitigated, then progress could be made on mode confusion.  

A standard approach to mitigating mode confusion is to employ ``autonomy explanation engines'' through user interfaces. However, our next use case argues that user interfaces are fundamentally limited by the underlying shared control logic.

\item Improved user interfaces: underlying the functionality of user interfaces (visualization engines, autonomy explanation engines, etc) is shared control technology.  That is, the explanatory power of the UI derives from the logic that the shared control employs.  For this reason, a deficient shared control engine limits the capability of the UI (no matter how beautiful the visualization or how realistic the voice).   Consider Figure~\ref{fig:two-mode-autonomy}. In this situation, linear blending based shared control strategies are forced to do one of the following: 1) override the operator, 2) cede complete control to the operator (stop assisting the operator), or 3) take two safe inputs and generate an unsafe course of action.  No matter what UI/explanatory engine is used, the human operator will potentially end up confused, because the logic of linear blending is impoverished.  In contrast, the decision made by GSC is naturally intuitive, since it balances the statistics of the operator and the autonomy---thus making the job of the UI vastly easier.

\item Assistive wheelchairs in crowds: As described in~\cite{ian-wheelchair}, navigating a motorized wheelchair through human crowds is extremely difficult for Parkinson's patients; the complexity of such an environment as well as the dexterity required to handle a wheelchair joystick makes such situations compelling application spaces for assistive technologies.  However, as discussed in~\cite{trautman-smc-2015}, current incarnations of linear blending are exponentially likely to needlessly argue with Parkinson's operators in crowded environments.   GSC, on the other hand, is both a) designed to mitigate disagreement between the operator and the autonomy (a critical feature for this population) and b) is an extension of the work in~\cite{trautman-ijrr-2015}, which was shown to be the most safe and efficient fully autonomous crowd navigation algorithm in existence.

\item Assistance without knowledge of goal state: in many situations, global location of the autonomous platform may be unknown, and so global destination cannot be communicated to the autonomy (e.g., assistive wheelchairs in crowds cannot perform localization because the dynamics of the environment is so severe; many platforms (self driving cars, UAVs) often find themselves in GPS denied environments).  Providing assistance in this situation requires reasoning through multimodality: without full knowledge of the operator's desired end state, the operator model acquires more predictive modes, and so the autonomy is forced to assist through multimodality (alternatively, the goal is incrementally revealed to the autonomy as the operator makes decisions).  

As an example with assistive wheelchairs: imagine that a wheelchair bound patient wishes to cross to the other side of a crowded room, but the wheelchair does not have a global map of the room (and so the patient cannot designate the goal).  However, the patient struggles to navigate safely through the crowd and requires incremental assistance; he can only indicate local directions as the wheelchair and patient move towards the goal.  Can the wheelchair move through the crowd while inferring the user's desired end state?  Early case studies indicate that GSC might be a promising framework for this type of scenario.

\end{enumerate}

\bibliographystyle{abbrv}
{\footnotesize
\bibliography{../GSCvsCSC}

\begin{thebibliography}{10}

\bibitem{carlson-smc-2012}
T.~Carlson and Y.~Demiris.
\newblock Collaborative control for a robotic wheelchair: Collaborative control
  for a robotic wheelchair: Evaluation of performance, attention and workload.
\newblock {\em SMC}, 2012.

\bibitem{draganrss2012}
A.~Dragan and S.~Srinivasa.
\newblock Formalizing assistive teleoperation.
\newblock {\em Robotics: Science and Systems}, 2012.

\bibitem{dragan-ijrr-2013}
A.~Dragan and S.~Srinivasa.
\newblock A policy blending formalism for shared control.
\newblock {\em International Journal of Robotics Research}, 2013.

\bibitem{inigo-blasco-isrrobotik-2014}
P.~Inigo-Blasco, F.~D. del Rio, S.~V. Diaz, and D.~C. Muniz.
\newblock The shared control dynamic window approach for non-holonomic
  semi-autonomous robots.
\newblock In {\em International Symposium on Robotics/Robotik}, 2014.

\bibitem{lopes-embs-2010}
A.~Lopes, U.~Nunes, and L.~Vaz.
\newblock Assisted navigation based on shared-control, using discrete and
  sparse human-machine interfaces.
\newblock In {\em IEEE Engineering in Medicine and Biology Society}, 2010.

\bibitem{ian-wheelchair}
I.~Mitchell, P.~Viswanathan, B.~Adhikari, E.~Rothfels, and A.~Mackworth.
\newblock Shared control policies for safe wheelchair navigation of elderly
  adults with cognitive and mobility impairments: Designing a wizard of oz
  study.
\newblock In {\em American Control Conference}, 2014.

\bibitem{peinado-icra-2011}
G.~Peinado, C.~Urdiales, J.~Peula, M.~Fdez-Carmona, R.~Annicchiarico,
  F.~Sandoval, and C.~Caltagirone.
\newblock Navigation skills based profiling for collaborative wheelchair
  control.
\newblock In {\em IEEE International Conference on Robotics and Automation},
  2011.

\bibitem{poncela-smc-2009}
A.~Poncela, C.~Urdiales, E.~Perez, and F.~Sandoval.
\newblock A new efficiency-weighted strategy for continuous human/robot
  cooperation in navigation.
\newblock {\em IEEE Transaction on Systems, Man, and Cybernetics---Part A:
  Systems and Humans}, 2009.

\bibitem{toussaintrss2012}
K.~Rawlik, M.~Toussaint, and S.~Vijayakumar.
\newblock On stochastic optimal control and reinforcement learning by
  approximate inference.
\newblock In {\em Robotics: Science and Systems}, 2012.

\bibitem{rushby-mode-confusion}
J.~Rushby.
\newblock Using model checking to help discover mode confusions and other
  automation surprises.
\newblock In {\em Reliability Engineering and System Safety}, 2002.

\bibitem{trautman-smc-2015}
P.~Trautman.
\newblock Assistive planning in complex, dynamic environments.
\newblock In {\em IEEE Systems, Man, and Cybernetics
  (http://arxiv.org/abs/1506.06784)}, 2015.

\bibitem{trautman-4-challenges}
P.~Trautman.
\newblock A unified approach to four basic challenges in shared autonomy.
\newblock Technical report, http://arxiv.org/abs/1508.01545, 2015.

\bibitem{trautman-gsc}
P.~Trautman.
\newblock Breaking the human-robot deadlock: Surpassing shared control
  performance limits with sparse human-robot interaction.
\newblock In {\em RSS}, 2017.

\bibitem{trautmaniros}
P.~Trautman and A.~Krause.
\newblock Unfreezing the robot: Navigation in dense interacting crowds.
\newblock In {\em IROS}, 2010.

\bibitem{trautmanicra2013}
P.~Trautman, J.~Ma, A.~Krause, and R.~M. Murray.
\newblock Robot navigation in dense crowds: the case for cooperation.
\newblock In {\em ICRA}, 2013.

\bibitem{trautman-ijrr-2015}
P.~Trautman, J.~Ma, R.~M. Murray, and A.~Krause.
\newblock Robot navigation in dense human crowds: Statistical models and
  experimental studies of human robot cooperation.
\newblock {\em IJRR}, 2015.

\bibitem{urdiales-nsre-2013}
C.~Urdiales, E.~P{\'e}rez, G.~Peinado, M.~Fdez-Carmona, J.~Peula,
  R.~Annicchiarico, F.~Sandoval, and C.~Caltagirone.
\newblock On the construction of a skill-based wheelchair navigation profile.
\newblock {\em IEEE Transactions on Neural Systems and Rehabilitation
  Engineering}, 2013.

\bibitem{urdiales-autonrobots-2011}
C.~Urdiales, J.~Peula, M.~Fdez-Carmona, C.~Barru{\'e}, E.~P{\'e}rez,
  I.~S{\'a}nchez-Tato, J.~del Toro, F.~Galluppi, U.~Cort{\'e}s,
  R.~Annichiaricco, C.~Caltagirone, and F.~Sandoval.
\newblock A new multi-criteria optimization strategy for shared control in
  wheelchair assisted navigation.
\newblock {\em Autonomous Robots}, 2011.

\bibitem{wang-adaptive-shared-control}
H.~Wang and X.~Liu.
\newblock Adaptive shared control for a novel mobile assistive robot.
\newblock {\em IEEE/ASME Transaction on Mechatronics}, 2014.

\bibitem{wang-ras-2005}
M.~Wang and J.~Liu.
\newblock Interactive control for internet-based mobile robot teleoperation.
\newblock {\em Robotics and Autonomous Systems}, 2005.

\bibitem{yu-autonrobots-2003}
H.~Yu, M.~Spenko, and S.~Dubowsky.
\newblock An adaptive shared control system for an intelligent mobility aid for
  the elderly.
\newblock {\em Autonomous Robots}, 2003.

\end{thebibliography}
}

\end{document}